\documentclass{article}

\usepackage{PRIMEarxiv}

\usepackage{epsf}
\usepackage{graphicx}
\usepackage{floatflt}
\usepackage{color}
\usepackage{comment}
\usepackage[utf8]{inputenc}
\usepackage{amssymb}
\usepackage{amsmath}
\usepackage{amsfonts,amssymb,amsthm}
\usepackage{mathtools}
\usepackage{commath}
\usepackage{amsfonts}
\usepackage{subcaption}
\usepackage{algorithm}
\usepackage{algpseudocode}
\usepackage{setspace}
\usepackage{ulem}
\usepackage{multirow}
\usepackage{adjustbox}
\usepackage{nomencl}
\usepackage{booktabs}
\usepackage{multirow}
\usepackage{caption} 

\usepackage[tableposition=above]{caption}

\makenomenclature

\usepackage{etoolbox}
\renewcommand\nomgroup[1]{%
  \item[\bfseries
  \ifstrequal{#1}{S}{State Estimation}{%
  \ifstrequal{#1}{C}{Communication Network}{%
  \ifstrequal{#1}{M}{Machine Learning}{}}}%
]}

\floatname{algorithm}{Procedure}

{}
{}
{}

\title{Cross-Layered Distributed Data-Driven Framework for Enhanced Smart Grid Cyber-Physical Security
\thanks{\textit{This paper is a preprint of a paper submitted to IET Smart Grid. If accepted, the
copy of record will be available at the IET Digital Library.} }
}

\author{
  Allen Starke, Keerthiraj Nagaraj, Cody Ruben, Nader Aljohani, Sheng Zou, \\
  \textbf{Arturo Bretas, Janise McNair, Alina Zare} \\
  Electrical and Computer Engineering \\
  University of Florida \\
  1064 Center Dr, Gainesville, Fl. 32611, U.S. \\
  \texttt{\{email\}allen1.starke@ufl.edu; k.nagaraj@ufl.edu; cruben31@ufl.edu;} \\
  \texttt{nzjohani@taibahu.edu.sa; shengzou@ufl.edu; arturo@ece.ufl.edu;}\\
  \texttt{mcnair@ece.ufl.edu; azare@ece.ufl.edu}
}

\begin{document}
\maketitle

\begin{abstract}
\textcolor{black}{Smart Grid (SG) research and development has drawn much attention from academia, industry and government due to the great impact it will have on society, economics and the environment. Securing the SG is a considerably significant challenge due the increased dependency on communication networks to assist in physical process control, exposing them to various cyber-threats. In addition to attacks that change measurement values using False Data Injection (FDI) techniques,  attacks on the communication network may disrupt the power system's real-time operation by intercepting messages, or by flooding the communication channels with unnecessary data. Addressing these attacks requires a cross-layer approach. In this paper a cross-layered strategy is presented, called Cross-Layer Ensemble CorrDet with Adaptive Statistics(CECD-AS), which integrates the detection of faulty SG measurement data as well as inconsistent network inter-arrival times and transmission delays for more reliable and accurate anomaly detection and attack interpretation. Numerical results show that CECD-AS can detect multiple False Data Injections, Denial of Service (DoS) and Man In The Middle (MITM) attacks with a high F1-score compared to current approaches that only use SG measurement data for detection such as the traditional physics-based State Estimation, Ensemble CorrDet with Adaptive Statistics strategy and other machine learning classification-based detection schemes.}

\end{abstract}


\nomenclature[S]{$\mathbf{z}_{SG}$}{Measurement Vector}
\nomenclature[S]{$\mathbf{x}$}{State Vector}
\nomenclature[S]{$\mathbf{e}$}{Measurement Error Vector}
\nomenclature[S]{$d$}{Number of Measurements}
\nomenclature[S]{$N$}{Number of States}
\nomenclature[S]{$h$}{Function of Measurements in terms of States}
\nomenclature[S]{$J$}{WLS-SE objective function}
\nomenclature[S]{$R$}{Covariance Matrix of Measurements}
\nomenclature[S]{$\sigma$}{Measurement Standard Deviation}
\nomenclature[S]{$\mathbf{x^*}$}{Current State Estimation}
\nomenclature[S]{$\mathbf{z}_{SG}^*$}{Current Measurement Estimation}
\nomenclature[S]{$H$}{Jacobian Matrix of $h$}
\nomenclature[S]{$\mathbf{r}$}{Residual of Measurements Vector}
\nomenclature[S]{$P$}{WLS-SE Projection Matrix}
\nomenclature[S]{$\mathbf{e_D}$}{Detectable Error Vector}
\nomenclature[S]{$\mathbf{e_U}$}{Undetectable Error Vector}
\nomenclature[S]{$II$}{Measurement Innovation Index}
\nomenclature[S]{$CME$}{Composed Measurement Error}
\nomenclature[S]{$p$}{Chi-Squared Probability}
\nomenclature[S]{$\chi^2$}{Chi-Squared Threshold}

\nomenclature[M]{$\mathbf{\mu}$}{Mean of normal samples}
\nomenclature[M]{$\mathbf{\Sigma}$}{Covariance of normal samples}
\nomenclature[M]{$\mathbf{\mu}_m$}{Mean of normal samples on $\phi_m$}
\nomenclature[M]{$\mathbf{\Sigma}_m$}{Covariance of normal samples on $\phi_m$}
\nomenclature[M]{$\tau$}{Threshold value to classify abnormal samples on $\Phi_R$}
\nomenclature[M]{$\tau_m$}{Threshold value to classify abnormal samples on $\phi_m$}
\nomenclature[M]{$T$}{Set of $\tau_m$}
\nomenclature[M]{$\alpha$}{Weight parameter}
\nomenclature[M]{$M$}{Number of buses/local  Cross-layer CorrDet detector}
\nomenclature[M]{$\mathbf{Z}$}{Training set}
\nomenclature[M]{$\mathbf{\hat{Z}}$}{Testing set}
\nomenclature[M]{$\mathbf{Y}$}{Label for training set}
\nomenclature[M]{$\mathbf{\hat{Y}}$}{label for testing set}
\nomenclature[M]{$\phi_m$}{Symbol for m-th local  Cross-layer CorrDet detector}
\nomenclature[M]{$\Phi_E$}{Symbol for Cross-layer Ensemble CorrDet detector}
\nomenclature[M]{$\Phi_R$}{Symbol for CorrDet detector}
\nomenclature[M]{$\mathbf{\delta}_{Z,m}$}{Squared Mahalanobis distance of all training samples with respect to $\phi_m$}
\nomenclature[M]{$\mathbf{\delta}_{z_{\hat{k}}}$}{Squared Mahalanobis distance of of k-th testing sample with respect to $\Phi_E$}
\nomenclature[M]{$d$}{Number of measurement or inter-arrival time or time delay values}
\nomenclature[M]{$\mathbf{z}_{SG}$}{Vector of all measurement values}
\nomenclature[M]{$\mathbf{z}_{IAT}$}{Vector of all inter-arrival time values}
\nomenclature[M]{$\mathbf{z}_{TD}$}{Vector of all time delay values}
\nomenclature[M]{$\mathbf{z}$}{Triplet, the complete set of triple elements $[\mathbf{z}_{SG}, \mathbf{z}_{IAT}, \mathbf{z}_{TD}]$}
\nomenclature[M]{$\mathbf{z}^{(c)}$}{Triplet element, $[\mathbf{z}_{SG}^{(c)}, \mathbf{z}_{IAT}^{(c)}, \mathbf{z}_{TD}^{(c)}]$}
\nomenclature[M]{$\mathbf{z}_m$}{Selected triplet element, a set of triple elements with respect to $\phi_m$}
\nomenclature[M]{$\eta$}{Magnitude parameter for threshold estimation}
\nomenclature[M]{$\beta$}{Window size for threshold update}
\nomenclature[M]{$\delta_{m}^{ECD}$}{Squared Mahalanobis distance of sample $\mathbf{z}$ with respect to the distribution of normal samples on $\phi_m$}
\nomenclature[M]{$\mu_{thr,m}$}{Mean of the Mahalanobis distance values of $\mathbf{z}_m$ of all normal samples in training data}
\nomenclature[M]{$\sigma_{thr,m}$}{Covariance of the Mahalanobis distance values of $\mathbf{z}_m$ of all normal samples in training data}

\nomenclature[C]{$p_{util}$}{Use of the queueing system (i.e. traffic intensity)}
\nomenclature[C]{$\mathbf{\lambda}$}{Packet arrival rate into system}
\nomenclature[C]{$\mathbf{\mu}$}{Packet service rate at each system}
\nomenclature[C]{$W$}{Total waiting time or transmission delays}
\nomenclature[C]{$n$}{Total number of packets}
\nomenclature[C]{$X_{t}, X_{i}$}{A series of datapoints ordered by time}
\nomenclature[C]{$L(\bullet)$}{Slowly varying function where $lim_{x \rightarrow \mathbf{\infty}}\frac{L(tx)}{L(x)} = 1$ for all $t > 0$}
\nomenclature[C]{$H$}{Hurst parameter used to measure the degree of long-range dependence}
\nomenclature[C]{$\mathbf{\beta}$}{Parameter used for measuring degree of LRD in autocorrelation functions}
\nomenclature[C]{$\mathbf{\phi}$}{Scaling function in autoregressive integrated moving average}
\nomenclature[C]{$\mathbf{\psi}$}{Mother wavelet function in autoregressive integrated moving average}
\nomenclature[C]{$\mathbf{\epsilon}_t$}{White noise distribution of $X_{t}$}
\nomenclature[C]{$\mathbf{\sigma}_{t},\:\mathbf{\sigma}_{t}^2$}{Standard deviation and variance of $X_{t}$}
\nomenclature[C]{$\overline{\rm G}_{\mathbf{\xi},\mathbf{\sigma}(\mathbf{\mu}}(x)$}{Standard cumulative distribution function of generalized pareto distribution}
\nomenclature[C]{$\mathbf{\xi}$}{The shape parameter in the generalized pareto distribution}
\nomenclature[C]{$\mathbf{\sigma}$}{The scale parameter in the generalized pareto distribution}
\nomenclature[C]{$IAT$}{Inter-arrival times}
\nomenclature[C]{$TD$}{Transmission delay}
\nomenclature[C]{$PC$}{packet count (i.e. traffic volume)}

\printnomenclature

\section{Introduction}
The future power grid, or Smart Grid (SG), has drawn much attention from academia, industry and government due to the significant impact it will have on society, economics and the environment. Next generation SG systems integrate control, communication and computation to achieve stability, efficiency and robustness of physical control processes. Network communication introduces several drawbacks and opportunities. First, it introduces an exposure to cyber-threats~\cite{faragSGSecurity2014}. Recently, the first confirmed cyber-attack-initiated blackout occurred in Ukraine and caused a power outage that affected 225,000 customers~\cite{ukraine,ukraine2}. Similar malware was found in several systems that operate in the US power grid~\cite{icscert2016,ukraine2}. In addition to being a subject of National Security, blackouts have huge economic impact. For instance, the estimated cost of the 2003 Northwest blackout ranges from 4 to 10 billion U.S. dollars in the United States, and 2.3 billion Canadian dollars in Ontario ~\cite{psotf2004}. This realization reinforced the critical need for research on cyber-related power grid vulnerabilities~\cite{morgan2016majorcyber,perez2016cyberattack}. Recent research literature addresses physical control process reliability, but research on cyber-physical security of SGs is still developing. 
\color{black}
On the other hand, communication networks provide an opportunity for cross-layer awareness. Typically, the classical power grid is protected by isolated and uncoordinated devices that provide ad-hoc solutions for each protection problem. The lack of cooperation between these tools leaves them vulnerable to distributed attacks. Consider power systems stabilizers (PSS),  which are located at synchronous generators to provide protection from small disturbances. If either a cyber-attack or other disturbance occurs at any stage of the data collection process, the PSS can potentially malfunction, compromising SG stability. A cross-layer approach can provide situational awareness, generate a more accurate response and increase system reliability and resiliency through network redundancy. Just as Stuxnet exploited a vulnerability in a set of programmable logic controllers (PLCs) that controlled
centrifuges for nuclear fuel processing~\cite{iran}, similar vulnerabilities could be exploited in PLCs that control automatic systems in grid-connected equipment.  

Current research on the cyber-security of the power grid focuses on a process called state estimation (SE)~\cite{handschin1975,bretas2011,bretas2018extension,Bretas201943,Bretas2013Convergence}. SE uses real-time measurements and static data about the system topology to (1) estimate the state of the system and (2) perform various monitoring applications \cite{monticelli1999state}. One of the main applications performed is Bad Data Analysis, which uses statistical tests to determine if any of the measurements on the system have an error. \textcolor{black}{Errors in measurements can come from a variety of sources like faulty meters or cyber-attacks.} A cyber-attack on the measurements themselves is called a False Data Injection (FDI) and is the most common form of cyber-attack considered in the literature. Current solutions for detecting FDI attacks tend to focus on modelling the behavior of the attack, then use load forecasting by training various machine learning methods to detect relatively small changes in load distribution or in the state variables of the power grid \cite{8909811,8587473,7526441}. While forecasting has been a very effective method to detect false data injection within the power grid, these methods do not consider multiple hackers injecting false data to conduct a coordinated attack. In addition, these solutions do not address detecting other forms of cyber attacks that can negatively impact the performance of the power grid. The work in~\cite{9165966} and~\cite{8610995} focuses on developing real-time solutions for detecting and defending against denial-of-service \textcolor{black}{(DoS)} attacks. In~\cite{9165966}, a dynamic differential system is used that models the changing states of a metering infrastructure during DoS attacks and in~\cite{8610995} an online storage facility is proposed for access to faster and scalable data analysis.\color{black} In~\cite{zhangDT,depaceDT,kushalDT} solutions are proposed to protect against negative impacts on power grid state estimation due to delays in the transmission of state measurements and control signals.\color{black}The work in~\cite{9020426} demonstrates that the lack of strong integrity and authentication checks in a power grid's network communication protocols can allow hackers easy access to the system and may cause detrimental effects to performance. 

\begin{figure*}[!t]
\centering
\includegraphics[height=8cm]{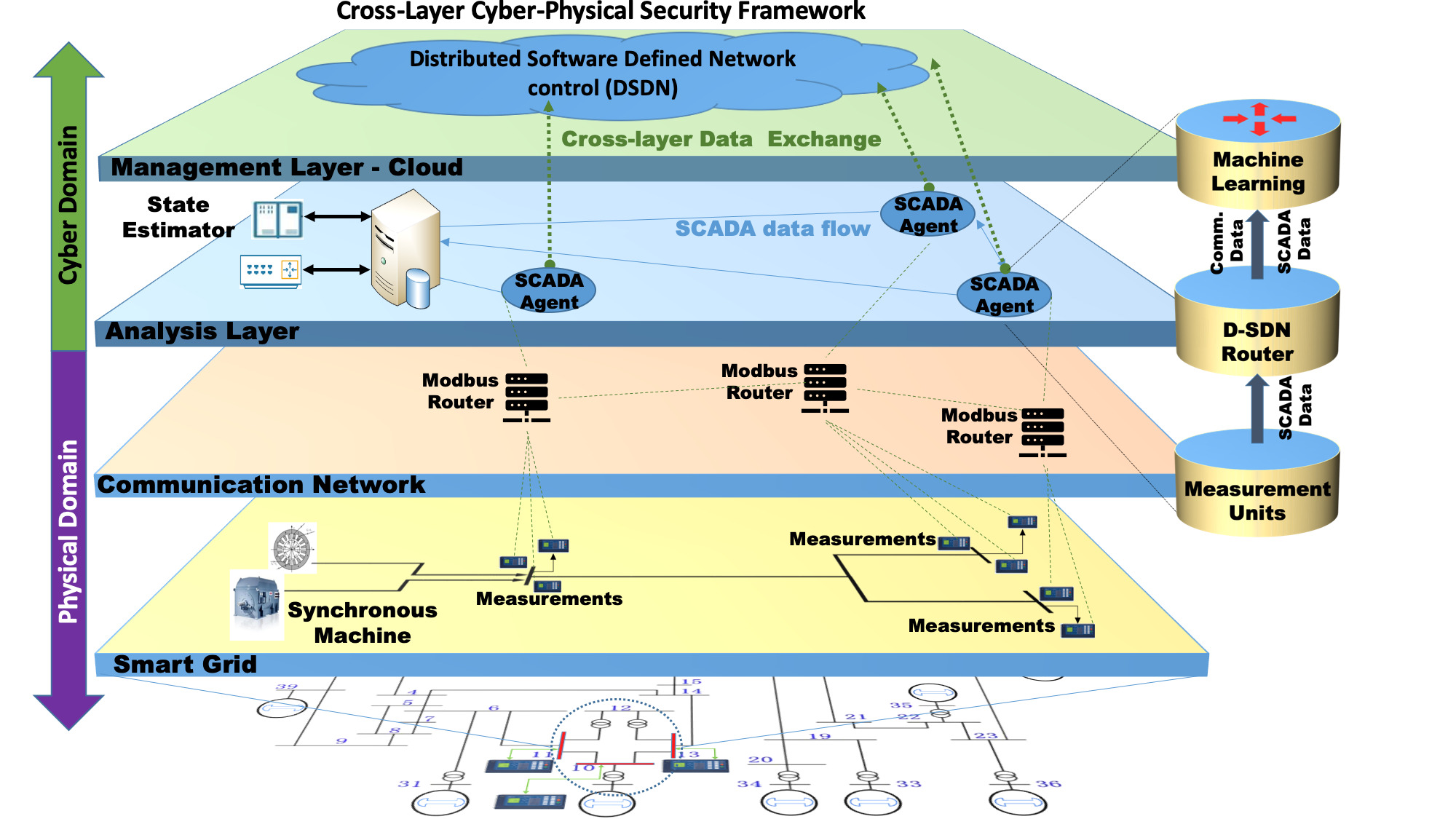}
\caption{Cross-layer Cyber-Physical Security Architecture}
\label{fig:cross}
\end{figure*}

 While there are a wide range of cyber attacks that can negatively impact the power grid, recent research has focused mainly on detecting and mitigating a single attack \textcolor{black}{, i.e., have assumed that only one type of attack occurs at a given time}. In a more realistic scenario cyber attacks involve multiple entities exploiting various security flaws in the physical and cyber domains of the cyber-physical system.  \textcolor{black}{To the best of our knowledge, the impact of a coordinated attack consisting of different types,  such as DoS, FDI, and Man in The Middle (MITM) has not been addressed. Furthermore, the challenge of leveraging the interdependence between different layers of the SG, i.e., physical domain and cyber domain, to achieve more robust securityin the system has not been sufficiently addressed.} Therefore, this paper proposes a method of detecting multiple attacks and different types of attacks, \textcolor{black}{initiated from different layers within the SG}. This is accomplished through our proposed, novel  cross-layer perspective. As shown in Fig.~\ref{fig:cross}, the SG cyber-physical system is composed of a physical domain, where measurements are taken and communicated through a communication network, and a cyber domain, where all of the data collected and communicated is analyzed. The cyber domain is where the SE process occurs. In a previous work~\cite{trevizan2019,ruben2019hybrid}, the authors have shown that Machine Learning (ML) can be used in the cyber domain, operating on the same data as the SE, to improve bad data analysis. This hybrid data-driven physics-model based framework takes advantage of both temporal data through ML and the known topology of the system through SE. Yet, this technique, like the other current research, still only addresses FDI attacks and only uses standard measurements taken on power systems. Thus, it does not consider the cross-layer interdependencies of the SG. In another previous work~\cite{nagaraj2020adaptive}, the Ensemble CorrDet with Adaptive Statistics (ECD-AS) strategy was developed by the authors to analyze measurement data and packet contents. ECD-AS is also a data driven method for the detection of FDI attacks and considers the changing state of the SG. The limitation of this method is it only uses measurement data, limiting its ability to detect cyber-attacks focused on the communication network layer of the SG. However, the work in this paper will leverage the analysis layer, which can also consider data related to the communication network that drives the SG, specifically, the packet inter-arrival times, transmission delay, and packet count. Considering this type of data will expand the model of an FDI attack in the cyber domain as well as reveal models of different types of cyber-attacks that would go undetected by current approaches in the literature.

In this paper, we present a SG cyber-physical security framework based on a cross-layer perspective that focuses on detection of various cyber attacks called the Cross-Layer Ensemble CorrDet with Adaptive Statistics (CECD-AS). Differently from previous approaches that propose SGs cyber-physical security, we consider the SG and the communication network as one and propose security solutions for the entire system. In such a system, the characteristics and security specifications of each layer should be considered in a cross-layer model to provide specific integrated countermeasures. Current state of the art approaches focus only on data from the SG like power and voltage measurements. Analyzing data related to the communication network provides further information that can and should be used to better detect not only FDI attacks, but a variety of other potential cyber-attacks that may not affect measurement values at all.

The contributions of this work to the state of the art are as follows:

\begin{itemize}
    \item Using a cross-layer perspective between the power grid and communication network to enhance the detection of cyber attacks on the Smart Grid.
    \item Expanding on the ECD-AS method of anomaly detection on the Smart Grid to include information from the communication network.
    \item The consideration of various types of cyber attacks, beyond just FDI attacks, that can only be detected by using a cross-layer perspective.
    \item Mathematically modelling the impact of various cyber attacks on the power grid and the communication network simultaneously.
\end{itemize}

The remainder of the paper is organized as follows. In Section \ref{sec:background}, background information on the various aspects of the cross-layer perspective is presented. The details of the CECD-AS algorithm are shown in Section \ref{sec:CECDAS}. The results of numerical tests used to evaluate the performance of this method are shown and discussed in Section \ref{sec:results}. Finally, Section \ref{sec:conc} presents the conclusions of this work.

\section{Background Information} \label{sec:background}

\subsection{Smart Grid System Communication Architecture}
Traditionally, SG communications consisted of very low rate exchange of serial data point-to-point between deterministic nodes. With the creation of Phaser Measurement Units (PMUs) and dynamic networking infrastructures, SG systems’ network traffic is increasingly coming from heterogeneous network structures experiencing different types of events, requiring more complex networking and control protocols. Managing these evolving networks using traditional network management schemes can increase the cost of network operation and maintenance and leave significant vulnerabilities in fault tolerance and security. Software Defined Networking (SDN) is a networking paradigm in which the forwarding hardware is decoupled from control decisions. The network intelligence is logically centralized in software-based controllers (the control plane), and network devices become simple packet forwarding devices (the data plane) that can be programmed via an open interface. SDNs help to assemble new services and infrastructure quickly to meet dynamically changing environment objectives. Furthermore, the software implementation of the control plane and the built-in data collection mechanisms are excellent tools to implement Machine Learning (ML) network control applications~\cite{machine133}. Extracting knowledge from data collection to understand and predict the state of the SG network will be crucial to implement security management in the SG. Two significant challenges are proposed to be addressed: real-time SDN machine learning systems, and real-time SG communications management through a distributed SDN architecture. As outlined in Fig. 1, the ML takes in the signal information from the physical layer as well as the network traffic information from the management layer and monitors the data for anomalies and signal corruption. The ML can leverage interactions between the subsystems on the physical layer and integrated operation with the Distributed Software Defined Networking (D-SDN) controllers at the management layer.

Traditional supervised ML methods, in general, assume that every possible class and distribution of possible samples for each of these classes are appropriately characterized by training data. Yet, in implementation, it is infeasible to assume that all possible behaviors can be identified and characterized in training data prior to implementation of a system – malicious attacks and their associated behaviors on the communication grid can and will be re-imagined and re-implemented.  Thus, a system is needed that can adapt to changes in communication behavior based on cross-layer information and can robustly detect and classify anomalous communication packets in real time. In the literature, ML methods haven't been used within a cross-layer security framework to monitor for malicious behavior.

\color{black}

\subsection{Network Performance Statistics} \label{sec:net_performance}
The cross-layered analysis framework is based on the IEEE 118-bus system emulating the commonly used Modbus RTU over TCP/IP protocol for traditional SG environments. This system follows the poisson traffic model, since transmission of packets occur in batches every 4 seconds ~\cite{jain_1146410}. Each bus is modelled after the M/M/c queue ~\cite{haviv_2015}, i.e., c$\geq$1, where the arrival of packets follow a poisson process and the service time of the queue follows an exponential distribution. The utiliziation (i.e. traffic intensity) of the systems is represented as

\begin{equation} \label{eq:util} 
p_{util} = \frac{\mathbf{\lambda}}{\mathbf{\mu}}
\end{equation}
where $\mathbf{\lambda}$ refers to the arrival rate of packets to the system\textcolor{black}{, }and $\mathbf{\mu}$ refers to the service time of packets in the system. The inter-arrival time (IAT) is the time between packet arrivals and is one of the network performance metrics used in our analysis, which  has an exponential distribution with parameter $\mathbb{\lambda}$. For t $\geq$ 0, the probability density function  is
\begin{equation} \label{eq:wait} 
f(t) = \mathbf{\lambda}e^{-\mathbf{\lambda}t}
\end{equation}
\textcolor{black}{Hence,} the average IAT is defined as 
 \begin{equation} \label{eq:wait} 
IAT = \frac{1}{\mathbf{\lambda}}
\end{equation}
The service time s has an exponential distribution with parameter $\mathbf{\mu}$\textcolor{black}{,} and the probability density function is
 \begin{equation} \label{eq:wait} 
g(s) = \mathbf{\mu}e^{-\mathbf{\mu}s} \textcolor{black}{,\quad  \forall s \geq 0}
\end{equation}
where $\frac{1}{\mathbf{\mu}}$ is the average service time of the system. Using Little's theorem, the total waiting time, in our case what we define as the transmission delays (TD), can be measured as

\begin{equation} \label{eq:wait} 
W = TD = \frac{1}{\mathbf{\mu} - \mathbf{\lambda}}
\end{equation}
Lastly, the so called "normal" distribution of network packet arrivals (i.e. non-attacked packets) into each system are determined by the probability of seeing a number of packet arrivals in a period from [0,T] .

\begin{equation} \label{eq:poisson} 
P(n\:arrivals\:in\:interval\:T) = \frac{(\mathbf{\lambda}T)^n e^{-\mathbf{\lambda}T}}{n!} 
\end{equation}
where T is the interval of time, and $n$ represents the number of packet. This is used to model traffic volume of the bus, and the packet count (PC) metric is defined as

\begin{equation} \label{eq:poisson} 
PC = \mathbf{\lambda}T
\end{equation}

\subsection{Cyber Attacks}
In this section, we provide an overview of the different possible types of cyber attacks based on the studies in~\cite{survery_testbed_7740849,survery_cps_7924372}, apply them to Supervisory Control and Data Acquisition (SCADA) networking systems for powergrid, and discuss their impact on network connectivity, topology and network traffic. 

\begin{itemize}
    \item False Data Injections (FDI): this type of cyber attack injects forged measurement into the control system in hope of misguiding the control algorithm. \newline
    \textit{Impact}: the impact factor of FDI attacks is high and can be seen on the physical and networking level of the victim system. The work ~\cite{tiny_os_7387331} proves that FDI attacks can increase the delay overtime of network packet transmissions. Such effects coupled with changes to power control have the potential of shutting down power grids causing blackouts for an entire region.  
    
    \item Flooding Denial of Service (DoS): in denial of service, the attacker intentionally disrupts the transmission of data to/from a given node through an excessive amount of service requests to the victim node, consuming all available resources. \newline
    \textit{Impact}: the impact factor of DoS attacks are high. This type of attack increases network traffic at its victim node (i.e. arrival rate) to consume the victim resources and extend queue length resulting in an increase in wait times or transmission delays as can be seen in \cite{icps_article}. This can cause nodes to shutdown, and impact the entire network as a whole. Distributed Denial of Service (DDoS) has a larger impact, since the attack occurs from multiple nodes resulting in a higher arrival/attack rate.
    
    \item Man-in-the-Middle (MITM): an attack in which a third party gains access to the communications between two other parties, without either of those parties realising it. The third party might read the contents of the communication, or in some cases also manipulate it. The attacker node advertises false information using Address Resolution Protocol (ARP) messages, or by hijacking the router each of the victim nodes are connected to. MITM attacks are seen as the grandfather to all other cyber attacks since this is typically the first step that must be achieved before executing other attacks. \newline
    \textit{Impact}: the impact factor of MITM attacks is typically low or non-existent. Based on the type of MITM attack a flux of packets could increase the network traffic with no serious impact on network performance. MITM attacks have the most potential out of all the attacks listed, since they are the hardest to detect and can evolve into a dangerous attack such as DoS or FDI.
\end{itemize}

\begin{figure}[t]
\centering
\includegraphics[width=\columnwidth]{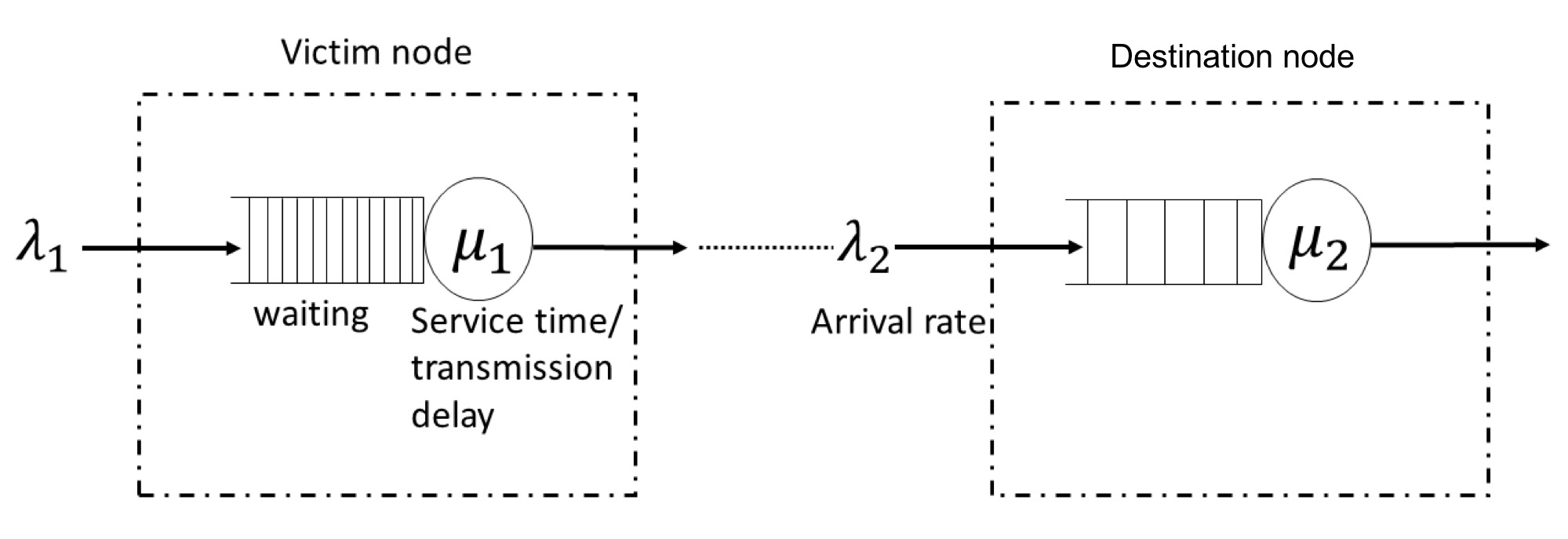}
\caption{Victim representation of a cyber attack}
\label{fig:mm1attack}
\end{figure}

\subsection{Cyber Attack Statistical Properties and Models} \label{sec:attack_models}
A cyber attack is an attempt by hackers to damage or destroy a computer network or system. In general, the cyber attacks discussed in the previous section demonstrate a similar characteristic of increasing the arrival rate of the network traffic at the victim or destination nodes, as demonstrated in Fig. ~\ref{fig:mm1attack}. During specific cyber attacks (e.g. DoS, FDI) the arrival rate, $\mathbf{\lambda}_1$, is a combination of normal and attack traffic (e.g. for DoS attacks $\mathbf{\lambda}_1$ is increased by flooded traffic from a hacker intending to crash the victim, where FDI attacks increase $\mathbf{\lambda}_1$ slightly to inject bad data). The impact of this increase of the arrival rate at the victim node (i.e. increase in $\mathbf{\lambda}_1$) is cascaded through the rest of the system, decreasing the service rate, $\mathbf{\mu}_1$, of the victim node and arrival rate, $\mathbf{\lambda}_2$, of the destination node. For MITM attacks the hacker infiltrates the connection between the victim node and the destination node, increasing the destination node's arrival rate, $\mathbf{\lambda}_2$, in an attempt to disguise themselves as the source node. Cyber attacks can be naturally modeled as stochastic processes~\cite{22277788}. The models are based on the attack rate (i.e. number of attacks that arrive per a unit of time), and can be represented on three levels of the networking system including network-level, victim-level, and port-level attack processes as stated in ~\cite{22277788}. FDI, DoS, and MITM attacks are executed on a victim-level attack process, which are attacks on individual victim computers or IP addresses. The authors of ~\cite{22277788} proved that 80\% and 70\% of cyber attacks on the network and victim level respectively don't function as Poisson processes and instead exhibit long-range dependence (LRD) with extreme values. LRD refers to the rate of decay of statistical dependence of two points with increasing time interval or spatial distance between the points. A phenomenon is usually considered to exhibit LRD if the dependence in its autocorrelation function (ACF) decays more slowly than an exponential decay.

\begin{equation} \label{eq:ACF} 
p(h) = Cor(X_{i},X_{i+h}) \: \mathbf{\sim}\: h^{-\mathbf{\beta}}L(h), h \rightarrow \mathbf{\infty}
\end{equation}

where $0 < \mathbf{\beta} < 1$ and $L(\mathbf{\bullet})$ is a slowly varying function meaning that $lim_{x \rightarrow \mathbf{\infty}}\frac{L(tx)}{L(x)} = 1$ for all $t > 0$ ~\cite{embrechts}. $\mathbf{\beta}$ is related to the Hurst parameter (H) as $\mathbf{\beta = 2-2H}$. H is used to measure the degree of LRD, where 1/2 $<$ H $<$ 1 and the degree of LRD increases as H $\rightarrow$ 1 ~\cite{22277788}.  The distribution of extreme values for stationary time series follows the generalized pareto distribution (GPD) with survival function

\begin{equation} \label{eq:GPD} 
\overline{\rm G}_{\mathbf{\xi},\mathbf{\sigma}(\mathbf{\mu)}}(x) = 1 - G_{\mathbf{\xi},\mathbf{\sigma}(\mathbf{\mu})} = \begin{cases}
(1+\mathbf{\xi}\frac{x}{\mathbf{\sigma}})^{-\frac{1}{\mathbf{\xi}}} & \quad \xi \ne 0, \\
e^{\frac{-x}{\mathbf{\sigma}}} & \quad \xi = 0.
\end{cases}
\end{equation}
where $\mathbf{\xi}$ and $\mathbf{\sigma}$ are called shape and scale parameter respectively. 

A combination of an (LRD)-aware model of auto regressive fractionally integrated moving average (ARFIMA or FARIMA) and extreme-value-aware model such as the integrated generalized autoregressive conditional heteroskedasticity (IGARCH) is used for modelling and predicting cyber attacks ~\cite{cryer_chan_2011}. FARIMA models are time series models that generalize ARIMA (autoregressive integrated moving average) models by allowing non-integer values of the differencing parameter. This is the well-known model where H = d + 1/2 and 0 $<$ d $<$ 1/2 . A stationary process $X_t$ is FARIMA (p,d,q) if 

\begin{equation} \label{eq:FARIMA} 
\mathbf{\phi}(B)(1-B)^{d}X_t = \mathbf{\psi}(B)\mathbf{\epsilon}_t
\end{equation}

for some -1/2 $<$ d $<$ 1/2, where 

\begin{equation} \label{eq:FARIMA2} 
\mathbf{\phi}(x) = 1 - \sum_{j=1}^p \mathbf{\phi}_j x^{j} 
\end{equation}

and 
\begin{equation} \label{eq:FARIMA2} 
 \mathbf{\psi}(x) = 1 + \sum_{j=1}^q \mathbf{\psi_j} x^{j} 
\end{equation}
B is the back shift operator defined by $BX_t = X_{t-1}, B^{2}X_t = X_{t-2}$, etc.~\cite{22277788}. GARCH is a model for identifying stochastic processes with a conditional variance of the process. A time series is a GARCH process if $X_{t}=\mathbf{\sigma}_{t}\mathbf{\epsilon}_{t}$ and the integrated GARCH model is as follows

\begin{equation} \label{eq:GARCH} 
\mathbf{\phi}(B)(1-B)\mathbf{\epsilon}_{t}^{2} = w + (1-\mathbf{\psi}(B))(\mathbf{\epsilon}_{t}^{2} - \mathbf{\sigma}_{t}^{2})
\end{equation}
where $\mathbf{\epsilon}_{t}$ is the white noise distribution, $\mathbf{\sigma}_{t}$ is the standard deviation, and $\mathbf{\sigma}_{t}^2$ is the variance~\cite{cryer_chan_2011}.

\color{black}

\subsection{Physics-Based State Estimation} \label{sec:psse}
In modern Energy Management Systems (EMS), the State Estimation (SE) process is the core process for situational awareness of a power system and is used in many EMS applications, including the detection of bad data.  The common approach to SE is using the classical Weighted Least Squares (WLS) method described in \cite{monticelli1999state}.  In this approach, the system is modeled as a set of non-linear equations based on the physics of the system:

\begin{equation} \mathbf{z}_{SG}=h(\mathbf{x})+\mathbf{e} \label{eq:SE}\end{equation}
where $\mathbf{z}_{SG}\in\mathbb{R}^{1 \times d}$ is the measurement vector, $\mathbf{x}\in\mathbb{R}^{1 \times N}$ is the vector of state variables, $h:\mathbb{R}^{1 \times N}\rightarrow\mathbb{R}^{1 \times d}$ is a continuously non-linear differentiable function, and $\mathbf{e}\in\mathbb{R}^{1 \times d}$ is the measurement error vector.  Each measurement error, $e_i$ is assumed to have zero mean, standard deviation $\sigma_i$ and Gaussian probability distribution.  $d$ is the number of measurements and $N$ is the number of states.

In the classical WLS approach, the best estimate of the state vector in \eqref{eq:SE} is found by minimizing the cost function $J(\mathbf{x})$:

\begin{equation} 
\label{eq:JSE} 
J(\mathbf{x})=\Vert \mathbf{z_{SG}}-h(\mathbf{x})\Vert _{R^{-1}}^{2}=[\mathbf{z}_{SG}-h(\mathbf{x})]^{T}R^{-1}[\mathbf{z}_{SG}-h(\mathbf{x})] 
\end{equation}
where $R$ is the covariance matrix of the measurements. In this paper, we consider the standard deviation of each measurement to be equal to 1\% of the measurement magnitude. In \cite{BRETAS2015484}, it is shown that in the gross error detection process all measurements should be weighted equally proportional to the measurement magnitude. After gross error processing, in the second step, meter precision can be restored and state estimation performed.

\begin{equation} \Delta \mathbf{z}_{SG}=H\Delta \mathbf{x}+\mathbf{e} \label{SElin} \end{equation}

where $H=\frac{\partial h}{\partial \mathbf{x}}$ is the Jacobian matrix of $h$ at the current state estimate $\mathbf{x}^*$, $\Delta \mathbf{z}_{SG}=\mathbf{z}_{SG}-h(\mathbf{x}^*)=\mathbf{z}_{SG}-\mathbf{z}_{SG}^*$ is the correction of the measurement vector and $\Delta \mathbf{x}=\mathbf{x}-\mathbf{x}^*$ is the correction of the state vector.  The WLS solution is the projection of $\Delta \mathbf{z}_{SG}$ onto the Jacobian space by a linear projection matrix $P$, i.e. $\Delta \mathbf{z}_{SG}=P\Delta\hat{\mathbf{z}}_{SG}$.  Letting $\mathbf{r}=\Delta \mathbf{z}_{SG}-\Delta\hat{\mathbf{z}}_{SG}$ be the residual vector, the $P$ matrix that minimizes $J(\mathbf{x})$ will be orthogonal to the Jacobian range space and to $\mathbf{r}$; $\Delta\hat{\mathbf{z}}=H\Delta\hat{\mathbf{x}}$.  This is in the form:
\begin{equation} \langle\Delta\hat{\mathbf{z}}_{SG},\mathbf{r}\rangle=(H\Delta\hat{\mathbf{x}})^{T}R^{-1}(\Delta \mathbf{z}_{SG}-H\Delta\hat{\mathbf{x}})=0.  \label{WLSsol}\end{equation}

Solving \eqref{WLSsol} for $\Delta\hat{\mathbf{x}}$:
\begin{equation} \Delta\hat{\mathbf{x}}=(H^{T}R^{-1}H)^{-1}H^{T}R^{-1}\Delta \mathbf{z}_{SG}. \label{dx} \end{equation}

At each iteration, a new incumbent solution $\mathbf{x}_{new}^*$ is found and updated following $\mathbf{x}_{new}^*= \mathbf{x}^*+\Delta\hat{\mathbf{x}}$. \eqref{dx} is solved each iteration until $\Delta \hat{\mathbf{x}}$ is sufficiently small to claim convergence of the solution.

The projection matrix $P$ is the idempotent matrix that has the following expression:
\begin{equation}\label{eq:P}
P = \Delta\hat{\mathbf{x}}=(H^{T}R^{-1}H)^{-1}H^{T}R^{-1}
\end{equation}

As shown in \cite{bretas2013geometrical}, the geometrical position of the measurement error in relation to the range space of $H$ provides another way of interpreting the state estimation. Hence, as the measurement vector can be decomposed into two subspaces, it is possible to decompose the measurement error vector into two components as follow:
\begin{equation} \label{eq:e}
\mathbf{e} = \underbrace{P\mathbf{e}}_{\mathbf{e_U}} + \underbrace{(I-P)\mathbf{e}}_{\mathbf{e_D}}
\end{equation}
The component $\mathbf{e_D}$ is the detectable error, which is the residual in the classical WLS model, while the component $\mathbf{e_U}$ is the undetectable error. $\mathbf{e_D}$ is in the orthogonal space to the range space of Jacobian whereas $\mathbf{e_U}$ is hidden in the Jacobian space.
\begin{equation} \label{eq:9}
\norm{\mathbf{e}}^{2} = \norm{\mathbf{e_D}}^{2} + \norm{\mathbf{e_U}}^{2}
\end{equation}
The error vector in (\ref{eq:9}) is called Composed Measurement Error ($CME$). In order to quantify the undetectable error, the Innovation Index ($II$) is introduced \cite{bretas2011innovation} and is presented in the following:
\begin{equation}\label{eq:10}
{II}_{i} = \frac{\norm{e^i_D}}{\norm{e^i_U}} = \frac{\sqrt{1-P_{ii}}}{\sqrt{P_{ii}}}
\end{equation}
Low Innovation index means there is a large component of error that is not reflected in the residual. Therefore, the residual will be very small even if there is a gross error. By using (\ref{eq:9})  and (\ref{eq:10}), the $CME$ can be expressed in terms of the residual and the innovation index as follow:
\begin{equation}\label{eq:11}
CME_i = r_i\left(\sqrt{1+\frac{1}{{II_i}^2}}\right).
\end{equation}

The $CME$ values for the measurements taken on the SG can then be used to do Bad Data Analysis, one of the main applications of SE \cite{bretas2017smart}. A chi-squared test is used for the detection of bad data in the measurement set, which compares a $CME$ based objective function value to a chi-squared threshold, which is based on the probability $p$ (typically $p=0.95$) and the degrees of freedom $d$:
\begin{equation} \label{eq:chi-squared}
J_{CME}(\hat{\mathbf{x}})=\sum_{i=1}^d \left[\frac{CME_i}{\sigma_i}\right]^2 > \chi^2_{d,p}
\end{equation}

If the value of $J_{CME}$ is greater than the chi-squared threshold, then an error is detected in the measurement set.

\subsection{CorrDet (CD) Anomaly Detection}
\label{corrdet}

The machine learning layer of the smart power grid uses the knowledge of already verified data to learn the normal state of a properly functioning grid. It is then able to detect any anomalies introduced into the system at any point forward and alerts the Network layer to identify the anomaly, isolate it from the remainder of the system and take appropriate action.
This action might be in the form of preventing contamination of the system, with regards to both power distribution in other subsystems, and data assimilation by the machine learning system itself.

The ML layer is developed using CorrDet Anomaly Detection \cite{ho2002linear, ho2000correlation, chang2002anomaly} algorithm as the foundation. The CorrDet anomaly detection learns a set of statistics for normal samples, including the mean ($\mathbf{\mu}$) and standard deviation ($\mathbf{\Sigma}$). Then, for each incoming sample ($\mathbf{z}$), the Mahalanobis distance of this new sample is computed with respect to the distribution of normal samples. A threshold value ($\tau$) is also estimated such that if the Mahalanobis distance is larger than the threshold value, the new sample is detected as abnormal sample.

There are two versions of CorrDet anomaly detection based on the change in statistics of data over time. If the data is not dynamically changing over time, a simple version of CorrDet algorithm can be applied. In this case, the statistics of normal samples are estimated directly from sample mean and covariance of training data and the threshold is picked by experiments. If the data is dynamically changing over time, an adaptive version of CorrDet algorithm can be used. In this case, the sample mean and covariance of training data, as well as the threshold value estimated from training data are only for initialization. As an incoming new samples is detected to be normal, the values of $\mathbf{\mu}$, $\mathbf{\Sigma}$ and $\tau$ are updated and the sample is added to the set of normal samples.
\newline

\section{Cyber-Attack Detection: A Cross Layered Perspective }\label{sec:CECDAS}

On the cyber domain, high-level functions take the input from data acquired at the physical layer and implement SE and ML actions that require interactions between the subsystems on the physical layer.
The top layer, as shown in Fig. \ref{fig:cross}, is the management layer where a D-SDN leverages data and statistics from communications and machine intelligence to detect abnormal behavior of the communication network and prompt the analysis layer to take corrective actions. The D-SDN output information is investigated for false data detection improvement, as well as detection of man-in-the-middle, and denial-of-service attacks. The D-SDN, through the embedded ML, will analyze network performance statistics (i.e. inter-arrival times, transmission delays, traffic volume, etc.) in the SG network communication layer. This network information is not considered by the SE physics-based model for detecting anomalous behavior.

\begin{figure}[ht]
\centering
\includegraphics[width=\columnwidth]{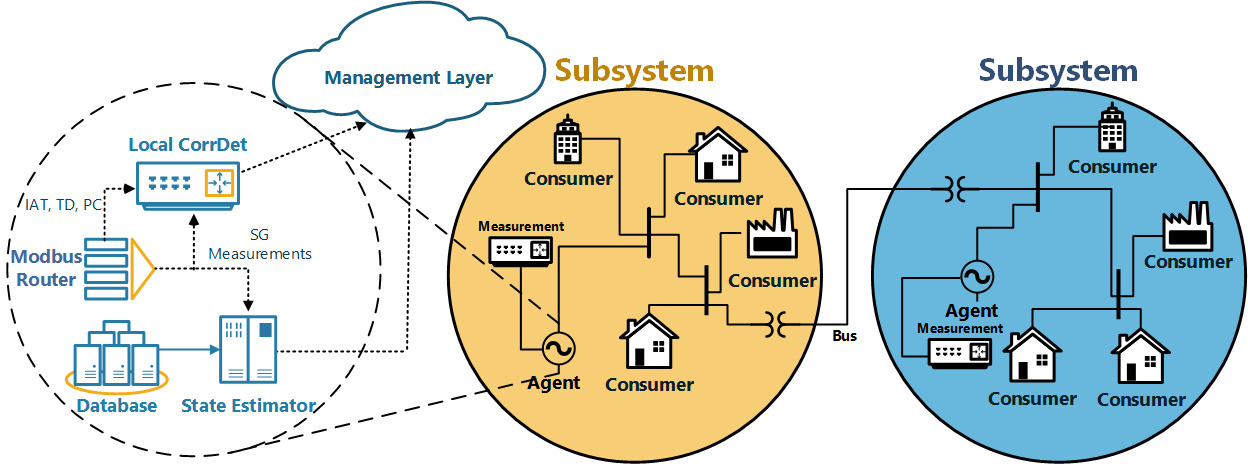}
\caption{Distributed architecture of the control and data acquisition systems of the Smart Grid}
\label{fig:dist-sub}
\end{figure}

Fig. \ref{fig:dist-sub} shows a more detailed view of the distributed SCADA Agents from Fig. \ref{fig:cross}. In each agent, a Modbus Router will collect real time measurements from the system and distribute that information to the SE as well as a Local  Cross-layer CorrDet. The Modbus Router also passes communication performance data to the Local  Cross-layer CorrDet, giving the ML algorithm the cross-layered perspective that is vital to detecting cyber-attacks besides FDIs. The Local Cross-layer CorrDets are then used in the Cross-layer Ensemble CorrDet with Adaptive Statistics algorithm described below. Finally, all of the analysis done through ML and SE are passed to the Management Layer.

\subsection{Cross-Layer Ensemble CorrDet with Adaptive Statistics (CECD-AS)}
\label{CECD-AS}

In the proposed anomaly detection using cross-layer data, each sample $\mathbf{z}$ is a concatenation of the measurement vector, $\mathbf{z}_{SG}$, inter-arrival time vector, $\mathbf{z}_{IAT}$, and transmission delay vector, $\mathbf{z}_{TD}$. In other words, each sample is a \textit{triple}, $\mathbf{z} = [\mathbf{z}_{SG}, \mathbf{z}_{IAT}, \mathbf{z}_{TD}]$, a $1 \times 3d$ vector instead of $\mathbf{z} = [\mathbf{z}_{SG}]$, a $1 \times d$ vector when only measurement values are used in FDI attack \cite{nagaraj2020adaptive}, where $d$ is the number of measurements/inter-arrival time/transmission delay of all buses in the power grid system. A \textit{triple element} $\mathbf{z}^{(c)}$, a $1 \times 3$ vector, is defined as a concatenation of the $c$th measurement value, the $c$th inter-arrival time value and the $c$th transmission delay value , $\mathbf{z}^{(c)} = [\mathbf{z}_{SG}^{(c)}, \mathbf{z}_{IAT}^{(c)}, \mathbf{z}_{TD}^{(c)}]$. A \textit{triple element} is a subset of the \textit{triple}, $\mathbf{z}^{(c)} \in \mathbf{z}$.

\begin{figure*}[!t]
\centering
\includegraphics[height=8cm]{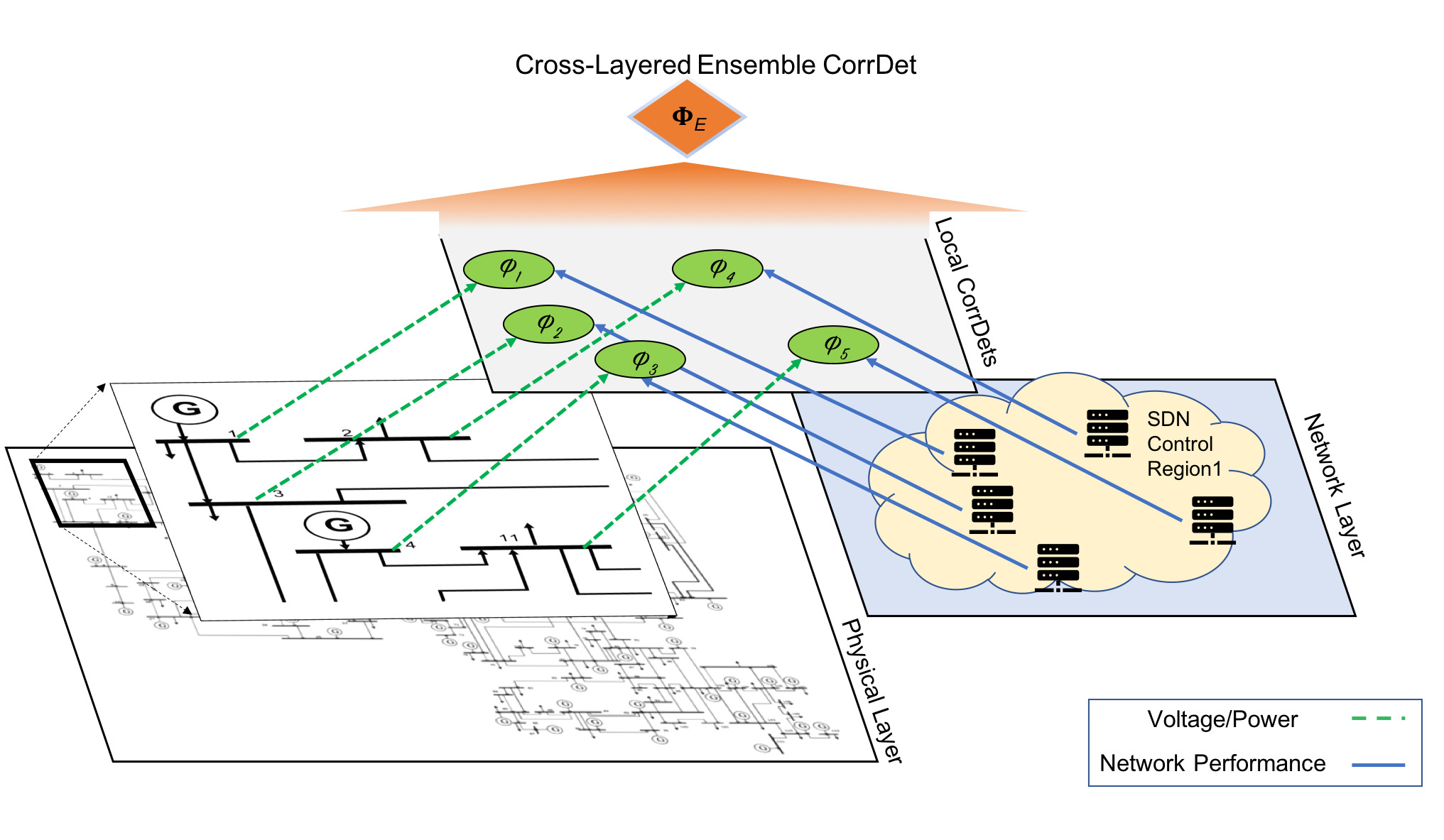}
\caption{Cross-Layer Ensemble CorrDet Framework}
\label{fig:ECD}
\end{figure*}

 The Cross-Layer Ensemble CorrDet with Adaptive Statistics, an extended work of CorrDet algorithm, can be considered as a set of Cross-layer CorrDet detectors for each local environment, as shown in Fig. \ref{fig:ECD}. For instance, local CorrDet for bus 1 ($\mathbf{\phi_{1}}$) receives measurement vector ($\mathbf{z}_{SG}$) from smart-grid layer, and inter-arrival time vector ($\mathbf{z}_{IAT}$) and transmission delay vector ($\mathbf{z}_{TD}$) from network layer for every sample. These local Cross-Layer CorrDets forms the Cross-Layer Ensemble CorrDet. In the whole power grid topology, spatially  neighboring  buses are more highly correlated and easier to be affected by an attack while buses that are further away have lower correlation. Thus, learning a full covariance over the \textit{triple} $\mathbf{z}$ of all buses is unnecessary (nearly sparse covariance), especially when training data is limited. Instead,  local,  fewer  dimensional  subsets of the \textit{triple}  offer  a  more accurate  statistic  estimation  and  a  computationally  cheaper,  more sensitive  anomaly  detection. 

The CorrDet detector learns a set of statistics, ($\mathbf{\mu}$, $\mathbf{\Sigma}$ and $\tau$) for all buses $\Phi_R$ in power grid topology, while CECD-AS learns a series of statistics ($\mathbf{\mu}_m$, $\mathbf{\Sigma}_m$ and $\tau_m$), one for each bus $\phi_m$ considering information from both smart grid and communication layers. CECD-AS detector prevents the numerical issue of estimating a high-dimensional mean and covariance for the distribution of the normal samples (in CorrDet detector) in the space of all \textit{triple elements} when the number of \textit{triple elements} are high and the number of training samples is low, by learning a lower-dimensional statistics in the space of only the measurements associated with each bus. 

There are $m_j (m_j<d)$ \textit{triple elements} on each bus $m$, where each bus is considered as a local, spatial region, corresponding to one Local Cross-layer CorrDet detector, $\phi_m$. For $\Phi_R$, the learning process consists of estimating $\mathbf{\mu}$ and $\mathbf{\Sigma}^{-1}$ from normal training samples $\mathbf{z}$ ($\mathbf{z} \in \mathbb{R}^{1 \times 3d}$). A similar strategy is proposed to learn the CECD-AS detector. The learning of $\Phi_E$ involves the estimation of a set of Local Cross-layer CorrDet detectors, $\phi_m$. 

For each $\phi_m$, the learning process consists of estimating its $\mathbf{\mu}_m$ and $\mathbf{\Sigma}^{-1}_m$ from the normal training samples with selected \textit{triple elements} $\mathbf{z}_{m}$ ($\mathbf{z}_{m}$ is a $1 \times 3m_j$ vector). The $\mathbf{\mu}_m$ and $\mathbf{\Sigma}^{-1}_m$ are initialized with the sample mean and covariance of selected \textit{triplet elements} of  first $k$ samples that are labeled as normal. 

Then, it starts to accept new sample and classify the new sample as follows. For each new incoming sample $\mathbf{z}$, a set of squared Mahalanobis distances, $\delta_{m}^{ECD}$, are computed using equation \eqref{ECorrDet-algorithm} and compared with the corresponding set of thresholds, $T$, where $T = \{\tau_m\}_{m=1:M}$. If at least one squared Mahalanobis distance in $\{\delta_{m}^{ECD}\}_{m=1
:M}$ is greater than its corresponding threshold, this incoming sample is classified as an anomaly. Otherwise, it is classified as a normal sample.

\begin{equation} \label{ECorrDet-algorithm} \delta_{m}^{ECD}(\mathbf{z}_{m})={(\mathbf{z}_{m}-\mathbf{\mu}_{m})}^T \Sigma_{m}^{-1} (\mathbf{z}_{m}-\mathbf{\mu}_{m}) \end{equation}

where $\mathbf{\mu}_m$ is the mean and $\Sigma_{m}^{-1}$ is the inverse covariance matrix of normal samples on $m$th local  Cross-layer CorrDet detector.

For each new sample that is classified as normal, the anomaly detector must be able to adapt with changing trends since data is dynamic and it changes gradually over time. Therefore, the mean, $\mathbf{\mu_m}$ and inverse covariance matrix, $\Sigma_m^{-1}$ are updated using the Woodbury Matrix Identity \cite{alveyzarecook2016} in equations \eqref{woodbury_mean_ECD} and \eqref{woodbury_cov_ECD} respectively. Note that this update is done only if the incoming data is considered normal data.

\begin{equation} \label{woodbury_mean_ECD} \mathbf{\mu}_{new,m} = (1-\alpha)\mathbf{\mu}_m + \alpha(\mathbf{z}_m-\mathbf{\mu}_m) \end{equation}

\begin{equation} \label{woodbury_cov_ECD} \Sigma^{-1}_{new, m} = \frac{1}{1-\alpha} \left[\Sigma^{-1}_m-\frac{(\mathbf{z}_m-\mathbf{\mu}_m)(\mathbf{z}_m-\mathbf{\mu}_m)^T}{\frac{1-\alpha}{\alpha}+(\mathbf{z}_m-\mathbf{\mu}_m)^T(\mathbf{z}_m-\mathbf{\mu}_m)}\right] \end{equation}

where $\mathbf{\mu}_m$ is the old mean of $m$th local Cross-layer CorrDet detector, $\Sigma^{-1}_m$ is the old inverse covariance matrix of $m$th local Cross-layer CorrDet detector and $\alpha$ is a hyper-parameter value between zero and one that determines how much importance is given to the new data sample versus the old mean. We determine the value of $k$ and $\alpha$ through experimentation.

The threshold can be assumed to be fixed for data set that has constant mean and small variation in the time domain. $T = \{\tau_m\}_{m=1:M}$ is estimated using equation \eqref{eq:threshold_noadapt}.

\begin{equation} \label{eq:threshold_noadapt} 
\mathbf{\tau}_m = \mathbf{\mu_{thr,m}} + \eta * \mathbf{\sigma_{thr,m}} 
\end{equation}

where $\mathbf{\mu_{thr,m}}$ and $\mathbf{\sigma_{thr,m}}$ are the mean and standard deviation of the Mahalanobis distance values of the selected triplet elements of all normal samples in training data. 

However, for the daily load profile data set considered in this work, the statistics of normal samples have a larger dynamically changing mean and covariance with time such that the previous fixed threshold assumption doesn't hold. Therefore, the Cross-Layer ECD with Adaptive Statistics is used in this work.

Unlike the fixed threshold estimation in CorrDet algorithm, adaptive threshold estimation in the CECD-AS algorithm not only initializes the threshold values $\tau_{m}$ for each local Cross-layer CorrDet detector (bus-level) following equation \eqref{eq:threshold_noadapt}, but updates $\tau_{m}$ in an online sliding window fashion \cite{nagaraj2020adaptive}.

 For every new incoming sample $\mathbf{z}$, the threshold values $\tau_{m}$ are inferred from the most recent $\beta$ normal samples before it. In other words, the standard deviation ($\sigma_{thr,m, -\beta}$) and mean ($\mathbf{\mu}_{thr,m, -\beta}$) of squared Mahalanobis distance values of $\beta$ normal samples past of the new sample $\mathbf{z}$ are calculated for each local Cross-layer CorrDet detector $\phi_m$. Here $\beta$ is the sliding window size. Then, threshold value $\tau_{m}$ for each local Cross-layer CorrDet detector is updated using equation \eqref{eq:adap_threshold} with updated $\mathbf{\mu}_{thr,m, -\beta}$ and $\sigma_{thr,m, -\beta}$, where $-\beta$ signifies the use of past $\beta$ number of samples for updating threshold.

\begin{equation} \label{eq:adap_threshold} 
\tau_{m} = \mu_{thr,m, -\beta} + \eta * \sigma_{thr,m, -\beta}.
\end{equation}

Let $K_1$ and $K_2$ be the number of training and testing samples, respectively. Let $\mathbf{Z}$ ($\mathbf{Z} \in \mathbb{R}^{{3d} \times K_1}$) and $\mathbf{\widetilde{Z}}$ ($\mathbf{\widetilde{Z}} \in \mathbb{R}^{{3d} \times K_2}$) the training and testing samples, respectively. Let $\mathbf{Y}$ ($\mathbf{Y} \in \mathbb{R}^{1 \times K_1}$) and $\mathbf{\widetilde{Y}}$ ($\mathbf{\widetilde{Y}} \in \mathbb{R}^{1 \times K_2}$) the corresponding labels. $\mathbf{\delta}_{Z,m}$ ($\mathbf{\delta}_{Z,m} \in \mathbb{R}^{1 \times K_1}$) denotes the squared Mahalanobis distances of all training samples with respect to $m$th Cross-layer CorrDet classifier, $\phi_m$. $\mathbf{\delta}_{\widetilde{z}_k}$ ($\mathbf{\delta}_{\widetilde{z}_k}\in \mathbb{R}^{1 \times M}$) denotes the squared Mahalanobis distances of $k^{th}$ testing sample with respect to all local Cross-layer CorrDet classifiers, $\Phi_E$. Let $\mathbf{B}$ be the squared Mahalanobis distances of all normal samples in the sliding window with a length of $\beta$ ($\mathbf{B} \in \mathbb{R}^{1 \times \beta}$). The pseudo code for the proposed anomaly detection algorithm is shown in Procedure \ref{algo::ecd_algorithm}.  

\begin{algorithm}
   \caption{Cross-Layer Ensemble CorrDet with Adaptive Statistics (CECD-AS) algorithm} \label{algo::ecd_algorithm}
    \begin{algorithmic}[1]
    	\State  \emph{ Train a Cross-Layer Ensemble CorrDet classifier}: 
      \Require $\mathbf{Z}$, $\mathbf{Y}$, $\mathbf{\widetilde{Z}}$
      \For {Every local  Cross-layer CorrDet classifier $m = 1:M$} 

      \State Initialize the mean $\mathbf{\mu}_m$ and covariance $\Sigma^{-1}_m$ of normal statistics using the sample mean and covariance of normal samples in the training set with selected \textit{triple elements} associated with $\phi_m$
      \State Initialize the squared Mahalanobis distance $\mathbf{\delta}_{Z,m}$ using equation \eqref{ECorrDet-algorithm}
      \State Initialize the threshold $\tau_m$ using equation \eqref{eq:threshold_noadapt}
            
      \EndFor
      \item[]
          	\State  \emph{Test using the Cross-Layer Ensemble CorrDet classifier with Adaptive Statistics}:
      \item[]
      \For {Every test sample $k = 1: K_2$}
      
      \State Compute the squared Mahalanobis distance $\mathbf{\delta}_{\widetilde{z}_k}$ using equation \eqref{ECorrDet-algorithm}
      
      \If {$\forall m,  \mathbf{\delta}_{\widetilde{z}_k} < \tau_m$}
      \State Classify $\widetilde{z}_k$ as normal sample: $\widetilde{y}_k = 0$
      \State Update the mean $\mathbf{\mu}_m$ and covariance $\Sigma^{-1}_m$ using equation \eqref{woodbury_mean_ECD} and equation \eqref{woodbury_cov_ECD}
      \State Update the sliding window by adding $\mathbf{\delta}_{\widetilde{z}_k}$ to $\mathbf{B}$ and removing the oldest value from $\mathbf{B}$.
      \State Update the mean $\mu_{thr,m, -\beta}$ and variance $\sigma_{thr,m, -\beta}$ of squared Mahalanobis distances in the updated sliding window of each local Cross-layer CorrDet detector 
      \State Update the threshold value $\tau_m$ for each local Cross-layer CorrDet detector using equation \eqref{eq:adap_threshold}
      \Else
      \State Classify $\widetilde{z}_k$ as abnormal sample: $\widetilde{y}_k = 1$
      \EndIf
      
     \EndFor
      \Ensure $\mathbf{\widetilde{Y}}$
\end{algorithmic}
\end{algorithm}

\subsection{Combining Power Grid and Communication Network Statistics}

Characteristics and security specifications of each layer should be considered in a cross-layer model to provide specific integrated countermeasures. The envisioned cross-layer analysis architecture for the cyber plane of one agent is represented in Fig. \ref{fig:dist-sub}. Each agent has a ML element with cross-layer interaction between the data plane, the SE, and the SDN to record and monitor network performance data and power exchanges. Packets are transmitted from bus-to-bus or bus-to-server using the Modbus RTU over TCP/IP networking protocol and technology. The transmission delays (TDs) are measured at the source node (i.e. $\frac{1}{\mathbf{\mu}_1 - \mathbf{\lambda}_1}$), as shown in Fig. \ref{fig:mm1attack}, and the inter arrival times (IATs) and traffic volume are measured at the destination node (i.e. $\frac{1}{\mathbf{\lambda}_2}$). Power measurements are polled and recorded periodically from each bus on the physical layer every 4 seconds. These values (i.e. voltage, power, and network performance statistics) are combined and analyzed locally using the algorithm described in section \ref{CECD-AS}, to provide a distributed approach at detecting anomalies for each agent. 

\color{black}
\section{Experimental results} \label{sec:results}
\begin{figure}[t]
\centering
\includegraphics[width=\columnwidth]{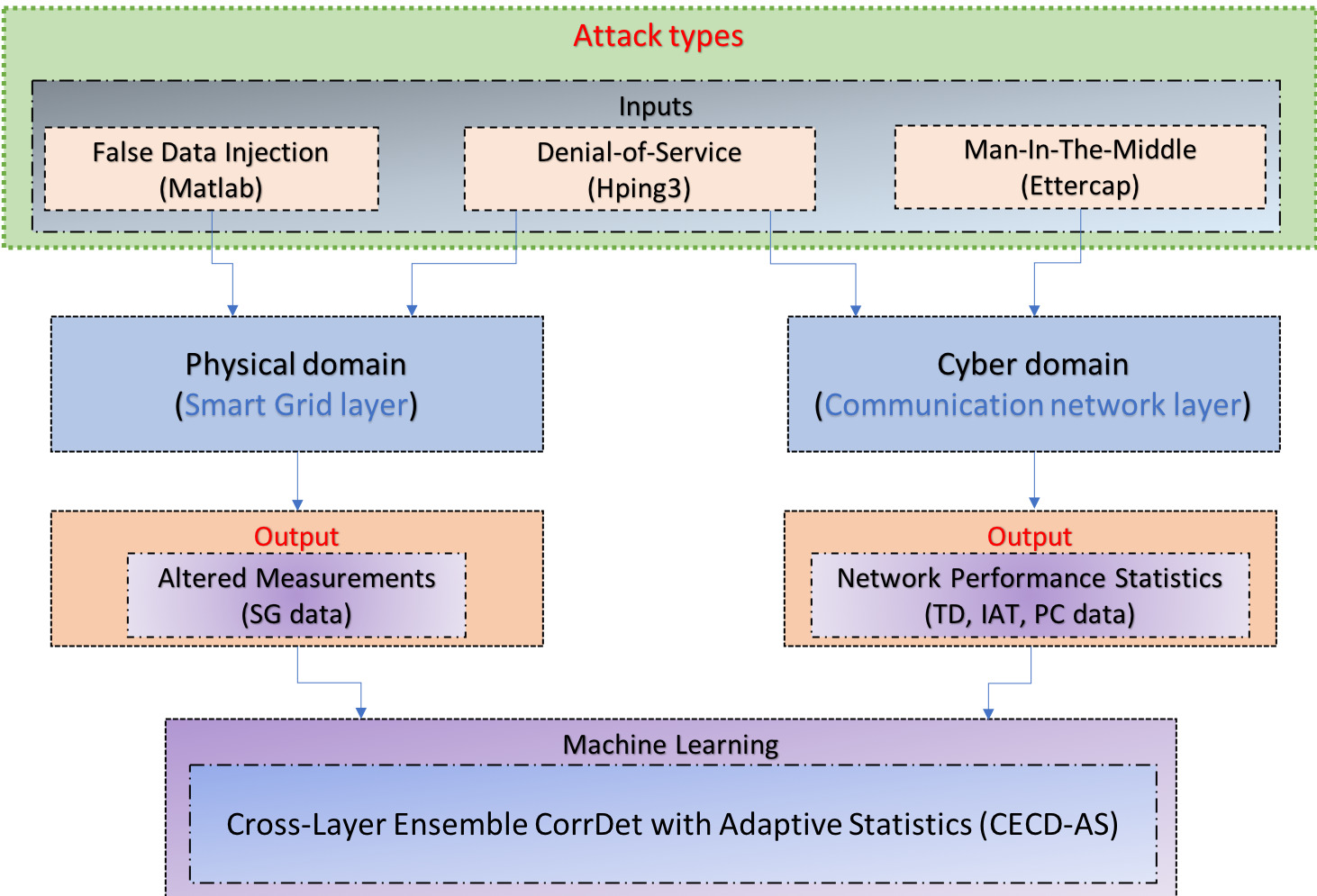}
\caption{\color{black}Simulation environment model.\color{black}}
\label{fig:sim_mod}
\end{figure}

The cross-layered analysis for detection of cyber attacks was validated using the IEEE 118-bus system. Using the MATLAB package MATPOWER \cite{matpower}, 21,600 samples (i.e. one day's worth) of measurement were generated with Gaussian noise  based  on  a  common  daily  load  profile that contains temporal information of a power system’s changing state. The measurement set included are real and reactive power flows, power injections, and all voltage magnitudes, resulting in 691 measurements. Then, network packet times (i.e. inter-arrival times, transmission delays) were generated using mininet and extensions to emulate the commonly used Modbus RTU over TCP/IP protocol for traditional SG environments ~\cite{fontes2015mininet}. Times were based on the M/M/c queue, i.e., $c \geq 1$,  where packet arrivals were modelled after Poisson distribution and transmission delays were inherently modelled after the exponential distribution, as described in Section ~\ref{sec:net_performance}. The correlating anomalous network performance statistics (i.e. IAT, and TD) from DoS and FDI attacks were generated based on the attack tools used in ~\cite{tiny_os_7387331,icps_article} respectively and exhibit the cyber attack properties described in Section ~\ref{sec:attack_models}. \color{black}The data flow of the simulation environment can be seen in Fig.~\ref{fig:sim_mod}. \color{black}
In general, the packet arrival rates to the victim node is increased during the periods where FDI and DoS attacks take place. The attack severity level is changed based on the different methods of attacking, as described in the next section, which can increase the network performance statistics by a factor between 2-7. Network traffic volume was generated using a combination of simulation tools such as mininet and a commonly used tool useful for executing man-in-the-middle cyber attacks called Ettercap ~\cite{ettercap_home_page}. For this paper, we tested our cross-layered anomaly detection scheme on three types of cyber attacks including multiple false data injections, multiple denial-of-service, and man-in-the-middle attacks. These cyber attacks are done on a victim-level attack process, which are attacks on individual victim computers or IP addresses.

\subsection{Data set description}

In this paper, four types of data sets were generated to simulate different types of attack. It should be noted that true labels are assigned during the process of introducing errors in all data set. \color{black} The datasets described used for experimentation are available upon request from any of the authors of this paper.\color{black}

\subsubsection{MFDI attack data set: }
 In this data set, only Multiple False Data Injection (MFDI) attack is considered. Errors are injected in the measurement set at random in 5\% of the data set samples. The resultant data set of this type has 1103 samples with MFDI attackout of 21600 samples. In this case, two scenario were considered. The first scenario, two measurements were chosen randomly to be compromised and included a higher severity level for the network statistics during the periods of the attack. In the second scenario, Coordinated Multiple FDI (C-MFDI) attack is considered. This attack is designed such that measurements with low Innovation Index (II) are attacked, which is shown to be difficult for physics-based state estimation solution bad data detection \cite{bretas2011}. The second scenario included a lower severity level for network statistics to make it more difficult in detecting the attack taking place.
 
\subsubsection{MDoS attack data set:}
In this data set, only Multiple Denial of Service (MDoS) attacks are considered. In a similar way, errors are introduced in 5\% of the samples at random. However, once a decision is made to introduce DoS attack, the measurements associated with the selected buses, which is also by random, will be altered for ten consecutive samples. The resultant data set of this type has 7330 samples with DoS attackout of 21600 samples. In each of the attacked samples, three buses were chosen randomly to have their associated measurements (voltage, power injection, and power flow from those buses) be compromised. The MDoS dataset contained a high severity level on the network statistics during the periods of attack. 

\subsubsection{MFDI-MDoS attack data set:}
In this data set, a combination of MFDI and MDoS at random samples with random number of measurements being in error is considered. The resultant data set of this type has 4861 samples with MFDI or MDoS attack out of 21600 samples. In each of the attacked samples, four measurements were chosen randomly to be compromised if the attack is MFDI. For MDoS attack, four buses at random were chosen to have their associated measurements be altered. The MFDI-MDoS dataset contained a high severity level on the network statistics during the periods of attack. 

\subsubsection{MITM attack data set:}
In this data set, only the man-in-the-middle (MITM) attack is considered. In a similar way to the MDoS data set, errors are introduced in 5\% of the samples at random. \textcolor{black}{Unlike DoS and FDI attacks, which affect both the power flow and network performance metrics, MITM attack only affects the network performance metric for traffic volume (i.e. packet count). Hence, traffic volume is generated and tested for this attack.} Once a decision is made to introduce the MITM attack, the network traffic of a randomly selected bus is altered for ten consecutive samples. The resultant data set of this type has 7330 samples with MITM attacks out of 21600 samples.

\begin{table*}[ht!]
    \renewcommand{\arraystretch}{2}
    \caption{Performance comparison of various false data detection methodologies~\cite{nagaraj2020adaptive}}
    \captionsetup{justification=centering}
    \centering
    \begin{tabular}{|c|c|c|c|c|}
    \toprule 
     \textbf{Method} & {\textbf{Accuracy}}  & {\textbf{Precision}} & {\textbf{Recall}} & {\textbf{F1-score}}    \\
       
        & \textbf{$\mu_{cv}$} $\pm$ \textbf{$\sigma_{cv}$} & \textbf{$\mu_{cv}$} $\pm$ \textbf{$\sigma_{cv}$} & \textbf{$\mu_{cv}$} $\pm$ \textbf{$\sigma_{cv}$} & \textbf{$\mu_{cv}$} $\pm$ \textbf{$\sigma_{cv}$} \\
        \midrule 
        
        KNN & 93.05 $\pm$ 02.71 & 13.36 $\pm$ 09.24 & 03.94 $\pm$ 03.35 & 04.64 $\pm$ 02.97 \\
        MLPNN & 78.04 $\pm$ 20.64 & 08.75 $\pm$ 11.98 & 20.89 $\pm$ 24.21 & 05.29 $\pm$ 04.09   \\
        GNB & 49.81 $\pm$ 38.06 & 28.26 $\pm$ 38.50 & 51.35 $\pm$ 39.73 & 07.94 $\pm$ 04.47   \\
        ADT & 45.56 $\pm$ 22.03 & 05.66 $\pm$ 02.33 & 57.84 $\pm$ 23.87 & 09.54 $\pm$ 01.54   \\
        SVC & 55.68 $\pm$ 25.75 & 11.46 $\pm$ 12.14 & 60.57 $\pm$ 22.11 & 14.89 $\pm$ 06.75 \\
        
        SE \cite{bretas2018extension} & 94.07 $\pm$ 00.25 & 36.56 $\pm$ 02.15 & 80.64 $\pm$ 02.21 & 57.03 $\pm$ 01.92   \\
        CD \cite{trevizan2019} & 04.88 $\pm$ 00.24 & 04.88 $\pm$ 00.24 & 99.07 $\pm$ 00.00 & 09.31 $\pm$ 00.43   \\
        ECD \cite{ruben2019hybrid} & 97.28 $\pm$ 01.40 & 46.52 $\pm$ 28.03 & 44.08 $\pm$ 29.63 & 55.42 $\pm$ 27.71  \\
        ECD-AS \cite{nagaraj2020adaptive} & \textbf{99.35} $\pm$ 00.45 & \textbf{87.24} $\pm$ 09.30 & \textbf{86.94} $\pm$ 09.87 & \textbf{92.54} $\pm$ 05.74 \\
        \bottomrule 
    \end{tabular}

    \label{tab:comp}
    
\end{table*}
\normalsize

\color{black}

\begin{table*}[ht!]
    \centering
    \caption{Performance results for MFDI, MDoS and MITM attacks(ECD-AS: Ensemble CorrDet with Adaptive Statistics \cite{nagaraj2020adaptive}, CECD-AS : Cross-Layer Ensemble CorrDet with Adaptive Statistics, MFDI : Multiple False Data Injection attacks, C-MFDI : Coordinated Multiple False Data Injection attacks, MDoS : Multiple Denial of Service attacks, MITM : Man In The Middle attacks, SG : Smart Grid measurements, IAT : Inter Arrival Time, TD : Transmission Delay, PC : Packet count)}
    \captionsetup{justification=centering}
    \renewcommand{\arraystretch}{1.5}

    {\small
    \resizebox{\textwidth}{!}{\begin{tabular}{@{}|c|c|c|c|c|c|c|c|@{}}
    \toprule
       \multirow{2}{*}{\textbf{Attack type}} & \multirow{2}{*}{\textbf{Method}} & \multirow{2}{*}{\textbf{Layer}}& \multirow{2}{*}{\textbf{Information}} & {\textbf{Accuracy}}  & {\textbf{Precision}} & {\textbf{Recall}} & {\textbf{F1-score}} \\ 
       
        & & & & \textbf{$\mu_{cv}$} $\pm$ \textbf{$\sigma_{cv}$} & \textbf{$\mu_{cv}$} $\pm$ \textbf{$\sigma_{cv}$} & \textbf{$\mu_{cv}$} $\pm$ \textbf{$\sigma_{cv}$} & \textbf{$\mu_{cv}$} $\pm$ \textbf{$\sigma_{cv}$} \\ 
        \hline
        MITM & ECD-AS & network &PC & 92.50 $\pm$ 00.21 & 91.62 $\pm$ 00.26 & 86.43 $\pm$ 00.29 & \textbf{88.95 $\pm$ 00.23} \\ \midrule
        
        \multirow{4}{*}{MFDI} & \multirow{3}{*}{ECD-AS} & smart grid & SG  & 99.36 $\pm$ 00.27 & 99.99 $\pm$ 00.01 & 87.35 $\pm$ 5.10 & 93.16 $\pm$ 03.00\\ \cmidrule(l){3-8} 
        & &  \multirow{2}{*}{network} & IAT & 98.58 $\pm$ 00.10 & 77.85 $\pm$ 00.86 & 99.99 $\pm$ 00.01 & 87.54 $\pm$ 00.54   \\
        & & & TD & 99.16 $\pm$ 00.04 & 85.56 $\pm$ 00.75 & 99.99 $\pm$ 00.01 & 92.22 $\pm$ 00.43  \\ \cmidrule(l){2-8} 
        & CECD-AS & cross-layer & [SG, IAT, TD] & 99.96 $\pm$ 00.01 & 99.41 $\pm$ 00.31 & 99.89 $\pm$ 00.14 & \textbf{99.65 $\pm$ 00.16} \\
        \midrule

        \multirow{4}{*}{C-MFDI} & \multirow{3}{*}{ECD-AS} & smart grid & SG  & 96.58 $\pm$ 00.78 & 99.94 $\pm$ 00.15 & 32.15 $\pm$ 13.39 & 47.13 $\pm$ 15.08\\ \cmidrule(l){3-8} 
        & &  \multirow{2}{*}{network} & IAT & 97.19 $\pm$ 00.44 & 70.50 $\pm$ 02.93 & 76.02 $\pm$ 08.27 & 72.82 $\pm$ 05.32   \\
        & & & TD & 96.27 $\pm$ 00.45 & 61.12 $\pm$ 02.64 & 68.73 $\pm$ 08.47 & 64.60 $\pm$ 05.09  \\ \cmidrule(l){2-8} 
        & CECD-AS & cross-layer & [SG, IAT, TD] & 99.83 $\pm$ 00.06 & 97.04 $\pm$ 01.12 & 99.77 $\pm$ 00.16 & \textbf{98.38 $\pm$ 00.61} \\ \midrule
        
        \multirow{4}{*}{MDoS} & \multirow{3}{*}{ECD-AS} & smart grid  & SG & 52.63 $\pm$ 00.78 & 34.08 $\pm$ 01.03 & 37.06 $\pm$ 04.15 & 35.41 $\pm$ 02.23   \\\cmidrule(l){3-8} 
        & & \multirow{2}{*}{network} & IAT & 99.98 $\pm$ 00.01 & 99.96 $\pm$ 00.01 & 99.99 $\pm$ 0.01 & 99.98 $\pm$ 00.01   \\
        & & & TD & 94.83 $\pm$ 00.13 & 87.28 $\pm$ 00.35 & 99.99 $\pm$ 00.01 & 93.20 $\pm$ 00.20  \\\cmidrule(l){2-8} 
        & CECD-AS & cross-layer & [SG, IAT, TD] & 99.83 $\pm$ 00.06 & 99.71 $\pm$ 00.08 & 99.82 $\pm$ 00.17 & \textbf{99.76 $\pm$ 00.09} \\
        \midrule
        \multirow{4}{*}{MFDI-MDoS} & \multirow{3}{*}{ECD-AS} & smart grid   & SG & 69.58 $\pm$ 01.98 & 28.96 $\pm$ 01.12 & 24.84 $\pm$ 01.92 & 26.66 $\pm$ 00.81  \\\cmidrule(l){3-8} 
        & & \multirow{2}{*}{network} & IAT & 90.53 $\pm$ 00.22 & 70.16 $\pm$ 01.32 & 99.99 $\pm$ 00.01 & 82.46 $\pm$ 00.01\\
        & & & TD & 98.81 $\pm$ 00.03 & 94.94 $\pm$ 00.29 & 99.99 $\pm$ 00.01 & 97.40 $\pm$ 00.15\\\cmidrule(l){2-8} 
        & CECD-AS & cross-layer & [SG, IAT, TD] & 99.64 $\pm$ 00.06 & 98.47 $\pm$ 00.27 & 99.96 $\pm$ 00.06 & \textbf{99.21 $\pm$ 00.14} \\

        \bottomrule
        
        \hline
        
    \end{tabular}}
    \hspace{1cm}

    \label{tab:tECD}
    
    }
\end{table*}

\subsection{Performance Analysis}

To evaluate the performance of the anomaly detection strategies included in this paper, we make use of the following classification metrics \cite{metrics}. In our analysis, True Negatives (TN) refer to normal samples that are predicted as normal samples. True Positives (TP) refer to anomalous samples correctly predicted as anomalous. False Negatives (FN) refers to anomalous samples predicted to be normal, and False Positives (FP) refers to normal samples predicted to be anomalous. 

\textbf{Accuracy} is the ratio of correctly predicted samples to the total number of samples. Accuracy is a good performance metric when the class sizes are balanced in the data set. For data sets with imbalanced class size (which is true in our analysis), accuracy would not serve as a good performance metric. Hence, we include metrics such as Precision, Recall and F1-score, which provides a better measure of performance for anomaly detection strategy. \eqref{PM_accuracy} shows the formula to calculate overall accuracy of the model. 

\begin{equation} \label{PM_accuracy} Accuracy = 100 \times \frac{TP + TN}{TP + FP + TN + FN} \end{equation}

\textbf{Precision} is the ratio of number of correctly predicted normal samples to the overall predicted normal samples. Precision is an important metric when we want to minimize False Positives. \eqref{PM_precision} shows the formula for calculating Precision performance metric.

\begin{equation} \label{PM_precision} Precision = 100 \times \frac{TP}{TP + FP} \end{equation}

\textbf{Recall} (also called as Sensitivity) is the ratio of number of correctly predicted normal samples to number of true normal samples. If we want to minimize the False Negatives, high value of Recall is expected without precision being too low. \eqref{PM_recall} shows the formula for calculating Recall performance metric.

\begin{equation} \label{PM_recall} Recall = 100 \times \frac{TP}{TP + FN} \end{equation}

\textbf{F1-score} is the harmonic mean of Precision and Recall. It would be better to have a single performance metric that would consider both Precision and Recall, and which strikes a good balance between them. This metric is more useful than accuracy since we have uneven class distribution for normal samples and anomalous samples in our analysis. \eqref{PM_f1score} shows the formula for calculating F1-score performance metric.

\begin{equation} \label{PM_f1score} \text{F1-score} = 100 \times \frac{2*Recall*Precision}{Recall + Precision} \end{equation}

\subsection{Numerical Results and Discussions}

\color{black}
To compare our results with other techniques, we first provide Table~\ref{tab:comp}, which shows the mean and standard deviation of accuracy, precision, recall and F1-score values of various FDI detection techniques developed using traditional machine learning classification algorithms~\cite{nagaraj2020adaptive}. Table~\ref{tab:comp} demonstrated that, for FDI attacks in daily load profile based SG data, the ECD-AS technique outperforms traditional ML techniques as well as the physics-based State Estimator and the data driven technique. In Table~\ref{tab:tECD}, we compare ECD-AS with the proposed cross-layer CECD-AS approach. Accuracy, precision, recall and F1-score are shown for MFDI, MDoS and MITM attacks.

\color{black}The Cross-Layer Ensemble CorrDet with Adaptive Statistics (CECD-AS) algorithm is employed to detect MFDI, MDoS, MITM and MFDI-MDoS attacks using various attack data sets. Model parameters such as $\alpha$ in \eqref{woodbury_mean_ECD} and \eqref{woodbury_cov_ECD}, $\beta$ in \eqref{eq:adap_threshold} and $\eta$ in \eqref{eq:adap_threshold} were selected through experimentation by considering the values of F1-score. The optimal values used to generate results in Table \ref{tab:tECD} are $\alpha = 8e-5$ and $\beta = 90$. The value of $\eta$ is selected to be $11$ for MFDI and C-MFDI attacks and $7$ for MDoS, MFDI-MDoS and MITM attacks. From the $21600$ samples in each data set, in the first cross validation experiment samples $0-1800$ ($K_{1} = 1800$, $2$ hours worth of data) were used for initial model training and samples $1800-12600$ ($K_{2} = 10800$, $12$ hours worth of data) were used for subsequent model update and testing phase of CECD-AS. For the second cross validation experiment, we consider samples $1000-2800$ ($K_{1} = 1800$) for training and samples $2800-13600$ ($K_{2} = 10800$) for subsequent model update and testing phase and so on for a total of $10$ cross validation experiments. 

\color{black}The data-driven statistical ML approach CECD-AS learns the behaviour of normal samples and anomalous samples through model training and adapts over time to effectively detect various kinds of attacks in the cross-layered cyber-physical smart grid systems. This claim is supported by the high F1-scores for MFDI, MDoS and MFDI-MDoS attacks in Table \ref{tab:tECD}. It is important to note that for these data sets, three out of the four methods tested are using new information when compared to the state of the art (SG Information) \cite{nagaraj2020adaptive}. IAT and TD information is tested individually along with the full CECD-AS framework to show that the most comprehensive and efficient method for cyber-attack detection is to combine the SG, IAT, and TD data. CECD-AS detects MITM attacks using the Packet Count (PC) data set and the performance metrics are shown in Table \ref{tab:tECD}. MITM attacks can only be detected from the data collected at network communication layer as the attack does not impact any measurement values at the SG layer. We recorded values of Accuracy, Precision, Recall and F1-score for all the cross validation experiments and Table \ref{tab:tECD} shows the mean ($\mu_{cv}$) and standard deviation ($\sigma_{cv}$) values for each of these metrics, where high value of $\mu_{cv}$ and low value of $\sigma_{cv}$ is favorable. 

\begin{figure*}[t]
\centering
\includegraphics[height=4.25cm]{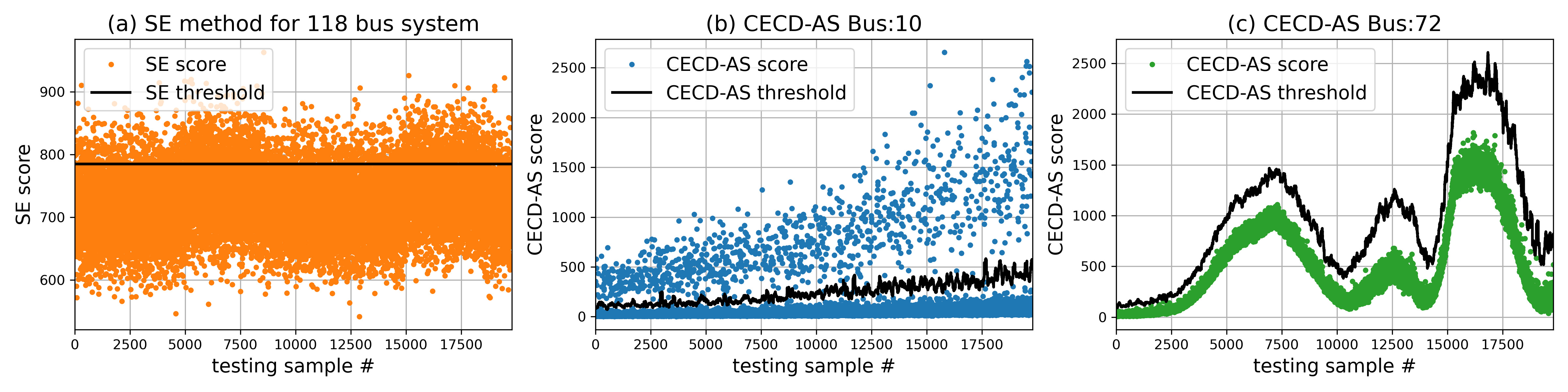}
\caption{Comparison of SE and CECD-AS methods through corresponding decision score and decision thresholds for Coordinated Multiple False Data Injection (C-MFDI) attacks}
\label{fig:se_cecd}
\end{figure*}

 Fig \ref{fig:se_cecd} shows the values of decision scores and thresholds obtained using SE and CECD-AS approaches for the C-MFDI dataset. SE approach only provides a single decision score for the entire 118-bus system for each testing sample, in oppose to the proposed CECD-AS which, due to its distributed nature, provides a decision score for each bus system. SE approach also results in a static decision threshold for the entire 118-bus system whereas CECD-AS approach results in adaptive threshold for each bus system. In the considered C-MFDI attacks dataset, bus 10 was targeted to inject false data. Fig \ref{fig:se_cecd}(b) shows the decision score and adaptive threshold obtained from CECD-AS approach for bus 10. As an illustration, we also show decision score and adaptive threshold for a bus that was not subjected to any C-MFDI attacks in Fig \ref{fig:se_cecd}(c). Decision scores which are above threshold are considered to be attacked samples and are compared with the ground truth to generate the performance results shown in Table \ref{tab:tECD}.\color{black}

\begin{figure}[h]
\centering
\includegraphics[width=\columnwidth]{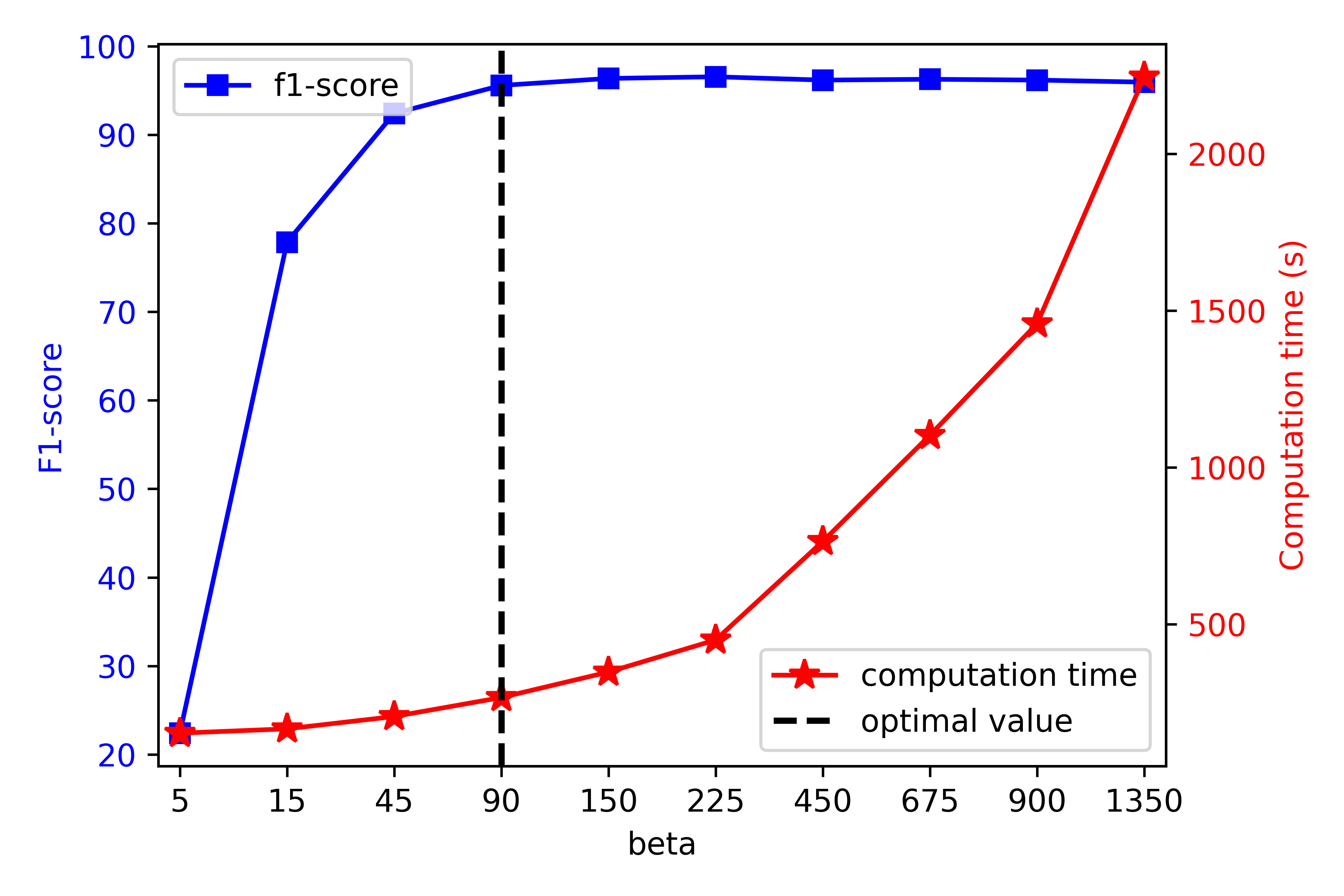}
\caption{Selection of optimal value for the beta based on F1-score and computation time}
\label{fig:betas}
\end{figure}

For every new incoming sample $\mathbf{z}$, the threshold values $\tau_{m}$ are inferred from the most recent $\beta$ normal samples before it. Smaller value of $\beta$ limits the number of past normal samples used to estimate anomaly threshold and would reduce the accuracy of detection as it would fail to capture the dynamically changing state in the data. Large value for $\beta$ would allow us to use more number of normal past samples but would add additional cost in storing and processing higher amount of data for every new incoming sample. This motivates us to pick an optimal value of $\beta$ through validation experiments, and we selected the value of $\beta$ considering both F1-score of detection on unseen data and computation time. In Fig \ref{fig:betas}, F1-score mostly increases with increasing beta but after beta reaches 90 the value of F1-score doesn't change by significant amount. Meanwhile computation time increases exponentially with increasing values of beta. We choose the value of beta with high f1-score and fairly low overall testing computation time as the optimal value.     

We can observe, from Table \ref{tab:physics}, the physics based SE approach \cite{bretas2017smart}  fails to consistently detect different types of attacks in the cyber-physical smart grid systems which is evident by the low F1-scores for MFDI, MDoS and MFDI-MDoS attacks. Most of the literature on SG bad data analysis research 0focuses on the recall metric, detecting as many of the FDI attacks as possible. Therefore, it makes sense the SE has a very high Recall score. However, there is a relatively high rate of FPs in the SE results, which when coupled with the fact that most samples are normal, leads to a low precision and F1 score, even when only analyzing MFDI attacks. For the data sets that include DoS attacks, the scores are bad all around since the DoS attack has little to no impact on actual measurement data. This means an attacker can be flooding the communication network with almost no detection by the SE. 

\begin{table}[ht!]
    \centering
    \renewcommand{\arraystretch}{1.5}
    \caption{Performance results for physics based SE \cite{bretas2017smart} (MFDI - Multiple \\ False Data Injection attacks, MDoS - Multiple Denial of Service attacks)}
    \captionsetup{justification=centering}
    \begin{tabular}{@{}|c|c|c|c|c|@{}}
    \toprule
        \textbf{Attack type} & \textbf{Accuracy} & \textbf{Precision} & \textbf{Recall} & \textbf{F1-score} \\ \hline
        MFDI            & 92.76    & 41.34     & 99.45   & 58.41    \\ \midrule
        MDoS            & 63.42    & 32.81     & 07.43   & 12.12    \\ \midrule
        MFDI-MDoS     & 75.18    & 38.37     & 16.95   & 23.51    \\ \bottomrule
        
    \end{tabular}
    
    \label{tab:physics}
\end{table}

For MFDI attacks, although accuracy values for individual data sets (SG, IAT, and TD) are close to accuracy value for cross-layered data set, F1-score gives a better picture of the model performance as it strikes better balance between precision and recall for the data sets with imbalanced class size. We can observe that cross-layered approach results in much higher mean F1-score ($99.65$) compared to the individual data sets. 

For C-MFDI attacks, ECD-AS approach fails to detect anomalies with high performance resulting in low F1-scores especially with SG measurements as SG layer is more affected by coordinated attacks on the grid compared to network layer. As the proposed CECD-AS takes information from both SG measurements and network layer data, it can capture most of the anomalies resulting in a mean F1-score of $98.38$. 

For MDoS attacks, the mean F1-score for SG data set is low due to the fact that MDoS attacks have little to no impact on the measurement values from SG layer, but higher impact on the parameters in network communication layer in the cyber-physical smart grid systems. Hence, CECD-AS results in high mean F1-scores for the data sets collected at network layer (IAT and TD). The cross-layered approach also results in similar performance as the individual data sets from the network communication layer.

For MFDI-MDoS attacks, the SG data set fails to give high mean F1-score due to the presence of MDoS attacks, and it is evident from Table \ref{tab:tECD} that the cross-layered approach results in higher mean F1-score ($99.21$). Performance metric values from Table \ref{tab:tECD} show that the proposed cross-layered approach can detect various kinds of attacks in cyber-physical SG systems with high F1-scores as the model will learn from the data sets obtained from both SG and network communications layers. IAT and TD data sets have resulted in higher recall values for different attack types compared to cross-layered approach but they also result in low precision values (high False Positives). Hence, results from Table \ref{tab:tECD} show that the proposed cross-layered approach results in both lower False Positives and False Negatives generally compared to results obtained from considering individual data sets from the SG layer or the network communication layer. 

\section{Conclusion}
\label{sec:conc}

This paper presents a cross-layered approach for enhancing the detection of different potential cyber-attacks on the Smart Grid. The state of the art solutions consider only false data injection (FDI) attacks as the source of influencing measurements data (voltages and powers). However, this paper showed that measurements can be affected through the use of communication network. Such an attack limits the ability of detecting bad data when considering measurement data alone. Hence, the novelty of the cross-layer strategy is the integration of data from communication network as another source of data beside measurements collected from Smart Grid in order to improve the anomaly detection scheme for real time monitoring. The Cross-Layer Ensemble CorrDet with Adaptive Statistics (CECD-AS) reflects the inter-dependency between the power grid and communication network.

The cross-layer strategy framework was implemented on the IEEE 118 bus system. Test results show that the proposed CECD-AS can detect attacks such as Multiple False Data Injections, Multiple Denial of Service and Man In The Middle with high F1-score compared to approaches which uses data from individual layers in the cyber-physical Smart Grid systems. The improved performance of the proposed framework can be attributed for its ability to understand normal and anomalous data behavior at both the Smart Grid layer (using voltage/power measurements) and network communication layer (using inter arrival times, transmission delay and traffic volume values) with a cross-layered perspective. The trained CECD-AS model decides whether a given sample is normal or anomalous and adapts its statistics in matter of milliseconds, so the proposed system can be implemented in real time monitoring of cyber-physical Smart Grid systems.

\section*{Acknowledgment}

This material is based upon work supported by the National Science Foundation under Grant Number 1809739.

\bibliographystyle{IEEEtran}
\bibliography{Main_Tex_CrossLayer_Paper_Clean}

\end{document}